# Language as an Independent Information Layer: A Conceptual Model of Communication, Cognition and Decision-Making

Anastasiia A. Alifanova[1[0009-0004-3019-1626]] and Elena N. Benderskaya[1[0000-0002-7733-416X]]

[1] Peter the Great St. Petersburg Polytechnic University, Polytechnicheskaya, 29, 195251 St.Petersburg, Russia
anastasiiaalifanova@gmail.com, helen.bend@gmail.com

**Abstract.** Based on an analysis of the role of language in thought and communication, this article proposes a new concept for designing corporate knowledge bases. The concept integrates the probabilistic vector space of a corporate vocabulary, reflecting industry specifics, subject focus, terminology, and culture, with traditional ontological modeling. This combination enables the efficient extraction of knowledge from accumulated corporate documents while strictly accounting for specific business processes. Consequently, this concept bridges statistical and semantic (cause-and-effect) methodologies. Furthermore, analyzing the projections of probabilistic spaces and causal relationships can help identify bottlenecks in business logic. As a dynamic system, language functions as a separate, independent layer within the overall information architecture. Introducing a dynamic component into the probabilistic space of word distribution allows it to be modeled as a multidimensional solution space for various problem formulations. In this context, input data defining the problem conditions serve as control parameters for dynamic transformations.

## 1 Introduction

The mere existence of language does not primarily distinguish humans from other biological species; mammals, birds, and even insects utilize communication systems to convey information. However, human language is uniquely set apart by distinct functional characteristics, including the capacity to exchange abstract information regarding the future or the past, structural universality across multiple domains beyond mere survival, and intentionality, which represents the conscious, willful control of speech rather than instinctual signaling.

Based on an analysis of these fundamental roles of language in thought and communication, this paper proposes a new approach to constructing enterprise knowledge bases. The method integrates a probabilistic vector space of corporate vocabulary, reflecting industry specifics, subject focus, terminology, and culture, with traditional ontological modeling. Consequently, statistical and semantic methodologies are bridged. Furthermore, analyzing the projections of probability spaces and causality locates the bottlenecks in business logic by extracting knowledge from accumulated enterprise documents while strictly accounting for specific business processes.

Noam Chomsky's research on human cognitive abilities [1] remains a foundational work on the role of language. It established key linguistic principles that later catalyzed developments in neurobiology, cognitive science, and machine learning. These principles encompass well-documented phenomena, such as the dependence of syntax on semantics and the human capacity for artificial language perception, alongside controversial areas like the biological origins of language and its function as a tool for thought rather than mere communication.

### 1.1 Motivation

The purpose of this work is to analyze research across neuroscience, cognitive science, linguistics (specifically semantics, syntax, and phonology), and machine learning to elucidate the role of language in communication and thought. Additionally, this study aims to determine the experimental limitations and inherent constraints of answering fundamental questions in this domain. Currently, Long Short-Term Memory (LSTM) models and correlation analysis techniques are employed to decode speech representations from neural data. This serves as a bridge between neuroscience and linguistics, as demonstrated by recent experiments in predicting neural time series [2].

The article is structured as follows. It begins by exploring the emergence of the problem and the evolution of perspectives driven by the development of new computational tools. Next, the paper examines the core functions of language and analyzes the relationship between syntactic and semantic processing within the framework of nonlinear dynamical systems theory. It then addresses the nature of thought, a question that has gained significant traction in the era of large language models. Finally, the study investigates how biosemiotics, alongside language, influences the thought process.

### 1.2 History of Research on the Relationship Between Language and Thought

The relationship between language, thought, and communication is intertwined with philosophy, linguistics, and the cognitive sciences. The hypotheses about the role of language emerged alongside the birth of philosophy. Each hypothesis was shaped by the specific methodologies and tools available during that era (see Tables 1, 2).

A modern foundation for understanding the connection between language and thought was established by Noam Chomsky's seminal book Syntactic Structures [1]. The central thesis was that grammar is biologically innate to the human species.

The emergence of cognitive science, an interdisciplinary field navigating the intersections of psychology, linguistics, philosophy, neuroscience, and computer science; allowed to practically validate these hypotheses in the end of 20$^{th}$ century. Cognitive linguistics developed as a subdiscipline dedicated to investigating human cognitive faculties and examining the relationship between linguistic structures and other forms of knowledge.

**Table 1.** Historical views on language and cognition.

| Period | Language for communication only | Language is connected to thought and communication |
|---|---|---|
| 400 BC | **Aristotle:** Language for communication only | **Plato:** Language and thought are linked |
| | - | **Stoics:** Language is inextricably linked to thought |
| 17-18 c. | **Rationalists:** Thinking does not depend on language | **Empiricists:** The question of the primacy of language or thought |
| 18-19 c. | **Leibniz:** Creation of a language of thought for the precise transmission of thoughts | **W. Humboldt:** Language influences the world's perception and is involved in thoughts' formation |

**Table 2.** Modern views on language and cognition.

| Period | Language as independent element | Language is connected to thought and communication |
|---|---|---|
| 1900-1920 | - | Research on bilingualism |
| 1920-1950 | Language is capable of forming consciousness, but this is not its primary task | **E. Sapir and B. Whorf:**<br>a) language completely determines thought;<br>b) and directs some cognitive categories |
| 1950-1980 | **Structuralism:** Language is a separate element | **N. Chomsky:** Grammar is innate in humans from birth |
| 21 c. | - | **Cognitive science:** language is connected with all cognitive and mental processes - memorization, retrieval of memories from memory, communication |
| | - | EEG, fMRI - provided Broca's and Wernicke's areas' functions |
| | - | a) language is responsible for planning, understanding the actions of others, memory;<br>b) language is connected with action |

While Broca's and Wernicke's areas were identified through lesion studies, the advent of EEG and fMRI prompted the broader functional dynamics. Modern research demonstrates that these regions are active not only during linguistic tasks but also during planning, action comprehension, and memory retrieval. Nevertheless, neuroimaging reveals that when an individual is exposed to action-oriented lexical stimuli, such as the verbs "run" or "squeeze," the specific cortical areas for executing those motor commands are activated without a physical action [3]. Psycholinguistic studies have validated hypotheses regarding parallel language activation in bilingual individuals, demonstrating that both linguistic systems remain simultaneously active even when the subject is communicating exclusively in a single language [4].

Research on bilingualism further elucidates these cognitive mechanisms [5]. For example, when the Russian word platye (dress) was played, bilingual subjects systematically fixed their gaze on a plate as well. Such studies confirm that language operates as the fundamental connecting link within the intersection of thought and communication.

### 1.3 Methods

The retrieved literature was identified, screened, and synthesized in accordance with the PRISMA 2020 statement, the core elements of which are reported here.

Studies were included if they addressed at least one of the three domains required by the research question: linguistic or theoretical accounts of the relationship between language and cognition; neurobiological evidence on the spatial or temporal organization of speech and meaning; or computational models of language, including conceptual frameworks, machine translation, ontological modelling, and LLMs. Eligible studies investigated languages from different language families, including Russian, English, and Chinese. Studies were excluded if they did not permit comparison across these domains or lacked identifiable methods or data.

Titles and abstracts were screened against the eligibility criteria, after which the full texts of potentially relevant records were assessed. Eighteen publications were retained as the evidence base for this article.

Included studies were compared along five dimensions: the object of analysis (syntax, semantics, articulation, cognition, or communication); the type of evidence (theoretical, neuroimaging, clinical, or computational); temporal or spatial resolution; the independence of language from cognition; and the implications for hybrid knowledge-base design. This critical comparative analysis serves as the basis for the conceptual model presented in Sect. 6.

## 2 A Contemporary Perspective on the Function of Language

Currently, no single theory dominates the field, and language is conceptualized both in conjunction with thought and as an independent faculty. For instance, based on an extensive analysis of neurobiological research spanning the past two decades, Fedorenko et al. [6] support the fundamentally communicative nature of language. Conversely, Borghi and Paglieri [7] insist on the capacity of language to modulate human cognitive functions. Meanwhile, Verwoert et al. [2] demonstrate that distinct components of speech, including articulation, acoustics, and semantics, are processed within the brain as partially independent parallel streams, while Zhu et al. [8] illustrate the precise spatiotemporal separation of syntactic and semantic processing.

### 2.1 Language as a Tool Relatively Independent of Thought

Reviews of the past two decades [6, 7] treat the language as a means of communication primarily. Supporting evidence includes pervasive of polysemy, the unlearnability of fully unambiguous artificial languages. At the same time, natural language can operate independently, because distinct aspects of language, including semantics, acoustics, and articulation, can be evaluated separately [1].

The stereo-electroencephalography (sEEG) identified distinct temporal windows of activation for articulatory, acoustic, and semantic representations, which demonstrate that linguistic form takes precedence over meaning within the neural processing chain. This is highly consistent with the view of language's relative autonomy.

### 2.2 The Relationship between Language, Communication, and Cognition

The fundamental role of language in the development of thought is supported by several critical lines of empirical evidence. To begin with, training children or non-human primates to utilize specific words or symbols denoting abstract concepts, particularly to highlight an environmental dimension relevant to a given task, has been shown to significantly improve their success in resolving complex problems that require relational reasoning [6, 7]. Furthermore, the acquisition of specific syntactic structures, such as object clauses, directly facilitates a child's ability to successfully perform Theory of Mind tasks, as robustly demonstrated by the behavioral experiments discussed in [6, 7].

At the same time, children who develop without early access to language [6] grow up with minimal or no linguistic input because they cannot perceive spoken language, and their caregivers do not know sign language. Although this prolonged lack of linguistic access has detrimental consequences for various aspects of cognitive development, such individuals consistently demonstrate the capacity for sophisticated cognitive operations: they can master mathematics, perform relational reasoning, construct complex causal chains and acquire a structured understanding of the world.

Furthermore, language plays a pivotal role in constructing social norms and cultural practices, which directly contributes to the development of higher-order cognitive skills and mutual understanding. This supports that the linguistic worldview is systematically formed through the pervasive influence of language on the conceptual and cognitive understanding of reality [9].

### 2.3 Methodological Contradictions in Empirical Evidence

Several genetic disorders, including Down syndrome and Williams syndrome, are characterized by varying degrees of intellectual disability, yet the linguistic abilities of individuals with these conditions remain remarkably close to typical development [8]. Similarly, certain neuropsychiatric disorders, such as schizophrenia, fundamentally impair the capacity for abstract thought and reasoning while leaving core language mechanics largely unaffected.

The studies cited above demonstrate the complex and non-linear nature of the relationship between language and thought, revealing that under specific conditions, the language network is activated prior to domain-general cognitive networks, whereas under alternative conditions, the reverse sequence occurs [6, 8]. This evident non-linearity allows the human thought process to be modeled as a trajectory of dynamic changes operating within a complex non-linear system.

## 3 Cognition as a Nonlinear Dynamics

### 3.1 Temporal Dynamics of Syntactic and Semantic Processing from a Neurobiological Perspective

Utilizing intracranial EEG (iEEG) data recorded from patients undergoing surgery for brain tumors, Zhu et al. [8] found that syntactic activity during to grammatical errors peaks at 200-40 ms, whereas semantic processing peaks at 400-800 milliseconds. Furthermore, the syntactic computation primarily activates the posterior inferior frontal gyrus, while semantic engages anterior regions. These findings support that the brain initially analyzes the structural, syntactic, and acoustic components of speech prior to correlating them with conceptual meaning.

A practical illustration can be observed during verbal interactions with children or novice language learners, when native speakers decipher the intended meaning by initially attempting to apply structural syntactic rules to the fragmented input, and subsequently determining the core message through the pragmatic context. From a neurobiological perspective, Verwoert et al. [2] highlighted the critical role of the superior temporal gyrus (STG) in mediating these processes: STG encodes intermediate, integrated acoustic-semantic representations that cannot be effectively processed in isolation by either the auditory or the semantic system alone.

### 3.2 Language as a Regulator of a Dynamic System

An example is semantic priming, whereby a specific term presented before an individual processes a subsequent text directly modulates its perceived meaning and shapes the spectrum of evoked associative networks. This demonstrates that an individual's linguistic experience not only systematically guides cognitive trajectories but also directly modulates valence and emotional evaluation.

As Zagumenov notes [10], reflection on language is not merely a means of conveying information but also a fundamental tool for thought and action, allows researchers to conceptualize language as a system possessing its own internal logic that directly influences the speaker's cognitive processes. However, the neurobiological mechanisms of language processing [2, 6, 8] operate in distinct cortical regions that are anatomically segregated from those dedicated to domain-general thought. Consequently, a profound contradiction arises regarding the exact modality through which language can function as a means of thought and action [10]. This presents a significant prospect for further research, which can be effectively resolved through the integration of linguistic, computational, and neurocognitive approaches.

Because certain thought processes remain fundamentally subconscious, researchers cannot easily isolate the primal cause of specific cognitive phenomena. Within this framework, language does not merely reflect the internal experiences, but actively reveals and governs their underlying cognitive mechanisms. Alongside the neurobiological approach, the intricate connection between language and thought can be productively evaluated from a biosemiotic perspective, which focuses on the evolutionary utility of sign systems. This allows language to be conceptualized as an adaptive tool that has developed over the course of biological and cultural evolution [11].

## 4 Language, Thinking, and Large Language Models

While the human thought process may be grounded in the operations of nonlinear dynamic systems, the question of whether cognition itself can be conceptualized as a purely probabilistic process emerged as LLM demonstrate a remarkable capacity to generate text despite lacking any neurobiological mechanisms. Borghi et al. [7] emphasize that fundamental architectural and mechanistic divergences separate artificial neural networks from human cognitive architectures, noting that the sophisticated text stems exclusively from immersive exposure to massive human-authored literature.

Recent research within the domains of computational linguistics and natural language processing [12] demonstrates that the expertise of linguists remains indispensable for complex tasks where calculating purely statistical probabilities is insufficient, particularly those requiring the verification of semantic coherence and the maintenance of structural linguistic cohesion.

The research on machine translation within the oil and gas industry [13] demonstrates that statistical algorithms fail in specialized tasks where a rigid ontological structure is necessary for establishing a baseline for model validation. At the same time, alphabetic and logographic languages remain fundamentally distinct in neurobiological terms, which is proven by the neuroimaging studies of Chinese-English bilingual individuals. The fMRI data demonstrated that processing these writing systems activates different neural pathways within the human brain [14].

By combining neurobiological data [2, 6, 8], cognitive-semiotic observations [9], and an analysis of LLM capabilities, we can conclude that language, thought, and communication form a nonlinear dynamic system with feedback loops. None of these components can be fully reduced to another; rather, each mutually influences the others. Ultimately, the thought process cannot be described in probabilistic terms, since such a hypothesis is based solely on the impressive results of LLMs.

## 5 A New Conceptual Framework for Business Analytics

If the human cognitive architecture operates as a complex nonlinear system [15], then language must be conceptualized not merely as a passive medium for data transmission, but as an active regulator that modulates the trajectories of this system. When designing intelligent business intelligence systems, modeling the human-machine interaction loop is a critically important element [15]. From this dynamical perspective, the probabilistic distribution of vocabulary establishes a multidimensional state space where mental representations continuously evolve. Linguistic elements, including semantic constraints and syntactic structures, function as control parameters that systematically alter the landscape of this space, effectively guiding cognitive processing toward stable attractors or opening new path-ways for problem-solving. This regulatory mechanism allows the system to remain highly adaptive, as shifting contextual parameters and novel inputs dynamically reshape the internal boundaries of the vocabulary space. Consequently, modeling language as a systemic regulator provides a robust theoretical foundation for translating biological cognitive dynamics into engineered information systems, allowing for the creation of enterprise knowledge bases that possess a high degree of structural flexibility and semantic resilience.

Thus, based on a detailed analysis of the role and functions of language in light of recent IT advancements and neurobiological findings on decision-making and information processing, we can conclude that modern business analytics systems require a new framework. This framework must account for language as an independent layer of information processing that actively shapes situation-appropriate decisions. The proposed concept integrates the statistical principle of data accumulation through embedding spaces with traditional ontological modeling, which serves as an intensional reflection of an organization's knowledge. Furthermore, the principle of dynamic development must be applied [16-18]; specifically, provision must be made for the continuous replenishment and modification of the mutual reflection between extensional and intensional structures, represented by their respective graphs. Methods for introducing this dynamic component should naturally be derived from dynamical systems theory, thereby implementing a multidisciplinary approach whose importance is widely recognized by contemporary researchers [19].

The proposed structural-functional architecture is formalized as a three-tier system exhibiting isomorphic mapping between cognitive dynamics and software engineering components (see Fig. 1). The development of the proposed approach is further supported by the thesis [20] stating that an effective solution to a complex problem should be based on both statistical data and theoretical constructs, implementing both «data-driven and theory-driven attitudes» at once [20].

To clarify the proposed framework, an 'independent information layer' must be distinguished from a system merely partially independent of human cognition. This complete independence is defined by three criteria: (1) autonomous structural logic, where semantic routing and graph topologies are governed by immanent laws rather than real-time cognitive states; (2) objectification and decoupling, meaning the layer exists as a separate digital substrate that preserves and accumulates data outside individual biological memory; and (3) nonlinear transformational capacity, which allows it to dynamically re-map, update, and generate emergent relationships internally. Consequently, while a partially independent system remains tethered to human cognitive cycles, the independent information layer operates as an external, self-sustaining symbolic engine.

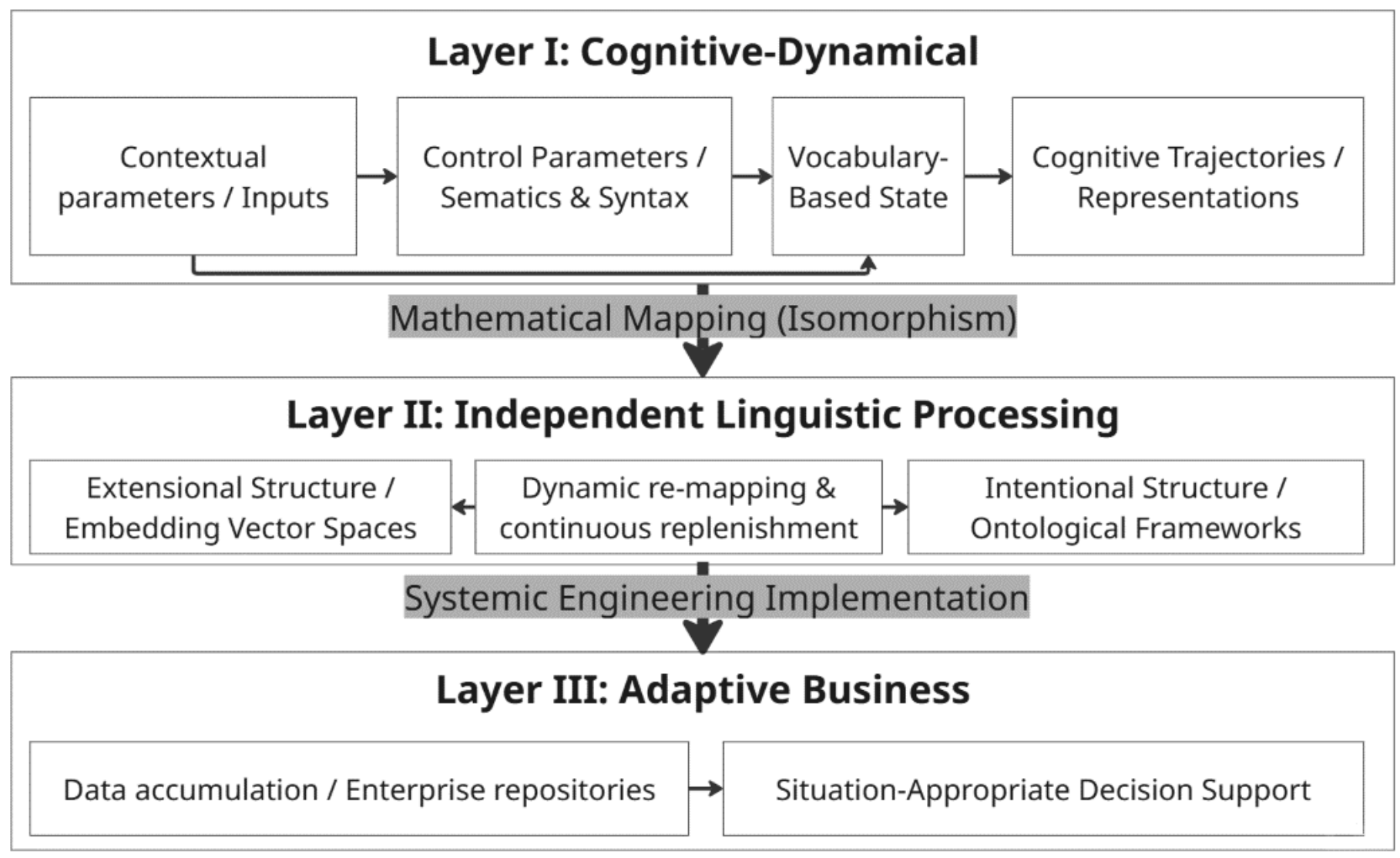

**Fig. 1.** The proposed multidisciplinary framework integrating dynamical systems theory into enterprise knowledge bases. Layer I represents the cognitive-dynamical abstraction; Layer II models the structural resilience through dynamic extensional-intensional graph re-mapping; Layer III outlines the system application in modern business analytics.

For the initial practical validation of the proposed three-layer architecture, a project-based enterprise dataset from knowledge-intensive industries (such as software engineering, aerospace, or industrial construction) would be most suitable. Specifically, the framework requires a corporate repository that aggregates retrospective project data, including 'Lessons Learned' logs, post-mortem reports, risk registers, and historical incident logs. This type of dataset is ideal because its unstructured and semi-structured text fragments contain explicit descriptions of past operational errors, strategic bottlenecks, and implemented solutions. Utilizing such data will allow us to test the architecture's capacity to extract semantic relationships from historical projects and dynamically re-map them to support decision-making in newly initiated organizational contexts.

## 6 Conclusions

In summary, the following conclusions can be drawn.

First, language, thought, and communication are three functionally distinct components. The existence of biological species capable of communicating and thinking without language confirms this theory. Nor does language serve as a mere link between thought and communication. If it did, it would be reasonable to assume that neither thought nor communication could exist without language—a premise that is not supported by empirical evidence.

Second, early humans did not possess a well-developed linguistic system, yet they successfully evolved and acquired the capacity for speech. This indicates that language emerged as an independent component only after human cognitive and communicative skills had already reached a relatively advanced stage.

Third, neuroimaging and clinical studies [2, 8] demonstrate that the brain regions responsible for the acoustic, articulatory, and semantic components of language are functionally distinct. These components are localized within partially independent neural networks and exhibit nonsynchronous activation patterns. This evidence confirms that language operates as an independent component functioning as a nonlinear dynamic system. Language should be understood not as the source of thought, but rather as an evolutionary tool that facilitates its development, enabling both the accumulation of knowledge and the articulation of complex mental constructs.

Fourth, while LLMs demonstrate the capacity to generate text by analyzing human-authored data, they still remain incapable of interpreting this information by a subjective, embodied experience.

Consequently, integrating the capabilities of statistical and knowledge-oriented approaches within a single framework governed by dynamic principles is of paramount importance. This hybridization should not merely combine disparate approaches within one system; rather, it must facilitate their joint application and mutual influence through the dynamic projections of the solution spaces they generate.

Future work will empirically validate the proposed dynamic graph re-mapping within a corporate knowledge base designed for cross-project knowledge transfer. To evaluate this framework, we will track two primary indicators: (1) the reduction in error repetition, a business-level metric quantifying the decline of recurring mistakes in newly initiated projects, and (2) the knowledge extraction rate, an analyst-performance metric measuring the ratio of relevant historical insights successfully retrieved and adapted for novel situations. Evaluating these parameters will demonstrate how dynamic graph re-mapping institutionalizes corporate experience into a reliable decision-support system.